\documentclass[sigconf]{acmart}
\AtBeginDocument{%
  }

\usepackage{float}
\usepackage{booktabs}
\usepackage{multirow}
\usepackage{makecell}
\usepackage{placeins}
\usepackage{hyperref}
\usepackage{subcaption}

\title{NeuroViz: Real-time Interactive Visualization of Forward and Backward Passes in Neural Network Training}

\author{Tanvi Sharma}
\affiliation{
  \institution{Boston University}
  \city{Boston}
  \state{MA}
  \country{USA}
}
\email{tsharma1@bu.edu}

\author{Reza Rawassizadeh}
\affiliation{
  \institution{Boston University}
  \city{Boston}
  \state{MA}
  \country{USA}
}
\email{reza@bu.edu}

\renewcommand\footnotetextcopyrightpermission[1]{}

\begin{document}

\begin{abstract}
Training neural networks is difficult to interpret, particularly for newcomers. We introduce \emph{NeuroViz}, an interactive visualization tool that supports real-time exploration of fully connected neural network training. Users can configure network architecture, activation functions, learning rates, and datasets, then observe activations, weight updates, and loss progression. \emph{NeuroViz} visualizes weight changes in direct correspondence with activation signals in both forward and backward passes, enabling users to distinguish pre- and post-update states within individual epochs and view dynamically updating per-neuron equations. We conduct a comparative user study with 31 participants against six established visualization tools and we achieved the highest usability score (SUS 80.97, in the `excellent' range), with mean rankings of 2.47 for clarity and 2.23 for usefulness (lower is better). Over 70\% of participants reported that the visualizations substantially increased their perception of neural network training transparency. The implemented instance is accessible at \href{https://neuroviz.org}{https://neuroviz.org}.

\end{abstract}

\begin{teaserfigure}
    \centering
    \includegraphics[
        width=0.7\textwidth,
        trim=0.5cm 0cm 1cm 0cm,
        clip
    ]{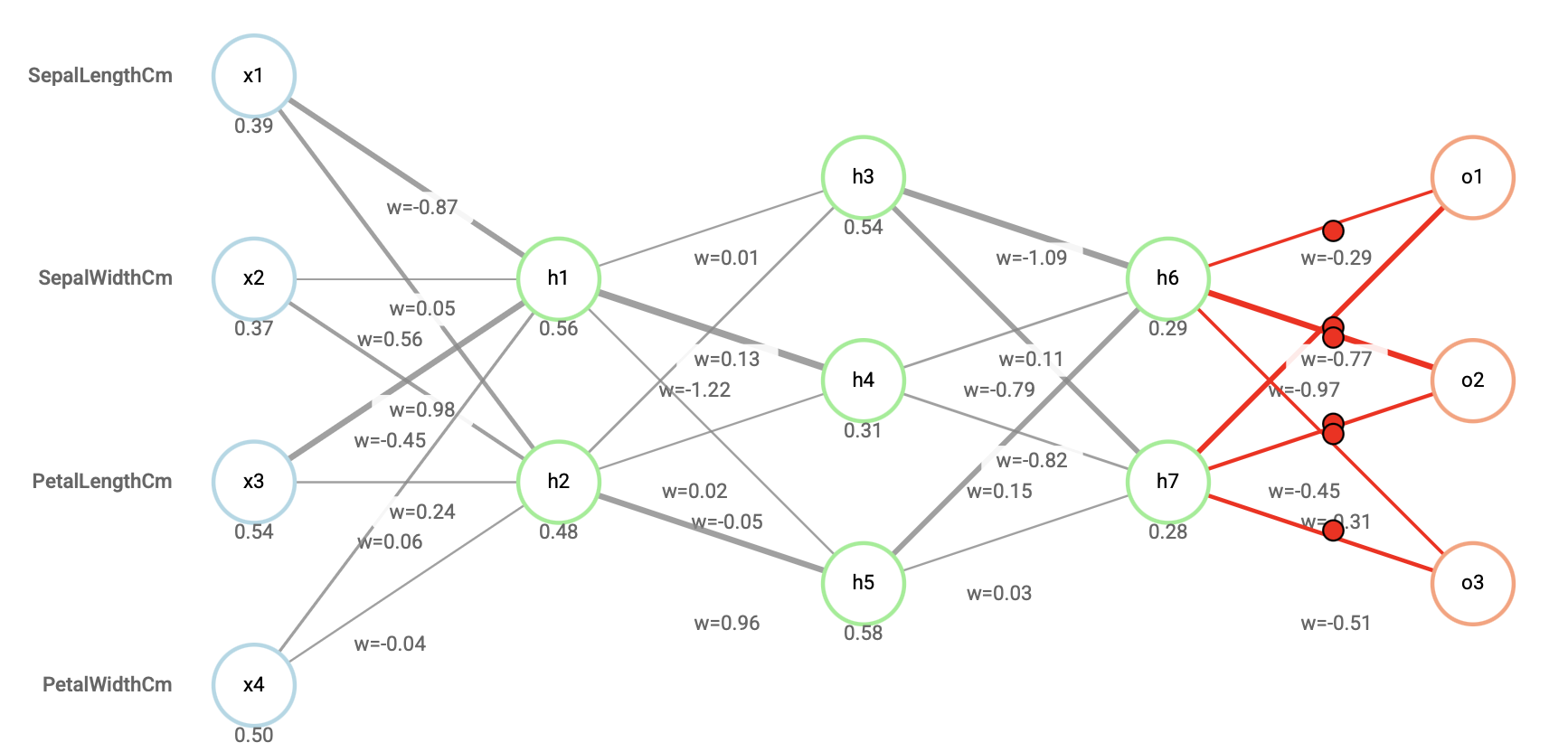}
    \vspace{-10pt}
    \caption{\emph{NeuroViz} visualizes the training on Iris dataset}
    \label{fig:NeuroViz_Image}
\end{teaserfigure}

\maketitle

\section{Introduction}

Neural networks have transformed machine learning, enabling major advances in computer vision~\cite{krizhevsky2012imagenet, he2016resnet}, natural language processing~\cite{vaswani2017attention, devlin2019bert}, and diverse domains~\cite{lecun2015deep, schmidhuber2015deep}. Yet their internal computations remain largely opaque~\cite{lipton2018mythos, castelvecchi2016blackbox}. Explainability is frequently cited as essential for building trustworthy AI systems~\cite{scienceDirect2024xai_survey, reza2025}, and comprehensive reviews emphasize that transparency is considered crucial for responsible deployment, especially in high-stakes settings such as healthcare and finance~\cite{yang2023xai_review}. 

This transparency creates practical challenges for students learning core concepts, instructors explaining training processes, practitioners debugging models, and researchers seeking to understand or trust model behavior~\cite{doshi2017towards, mittelstadt2019explaining}. Formal descriptions and code implementations accurately capture forward propagation, backpropagation, and gradient-based optimization~\cite{goodfellow2016deep, bishop2006pattern}, but they rarely convey the sequential, iterative nature of learning in an intuitive way~\cite{olah2018building, olah2020zoom}. Interactive visualization tools have thus become important aids for improving interpretability and human understanding of neural network behavior. 

Existing tools offer valuable insight. TensorFlow Playground~\cite{tfplayground} supports real-time experimentation with small feedforward networks. Netron~\cite{netron} provides static architecture rendering. CNN Explainer~\cite{cnnexplainer}, GAN Lab~\cite{ganlab}, LLM Visualization~\cite{bycroft_llm}, and Transformer Explainer~\cite{transformerexplainer} each target specific architectures or inference processes. However, these tools do not offer insight into the fine-grained timing of weight updates during training, particularly the relationship between signal propagation and parameter adjustment within individual epochs. Step-aligned depictions of backpropagation and weight update timing remain uncommon in interactive feedforward training tools.

We hypothesize that a visualization tool providing synchronized, step-aligned rendering of the complete training cycle, where forward activations, backward gradients, and weight updates are animated in direct temporal correspondence, can improve perceived interpretability, usability, and trust compared to existing approaches. To test our hypothesis, we introduce \emph{NeuroViz}, a web-based interactive tool for real-time exploration of feedforward neural network training. Users configure network architecture, activation functions, learning rates, and datasets, then observe activations, weight updates, and loss progression. Weight changes are visualized in direct correspondence with signal arrival at each neuron through pulse-based animations, enabling users to distinguish pre-update and post-update states within individual epochs. \emph{NeuroViz} also presents dynamically updating per-neuron equations and continuous performance metrics. 

We evaluated \emph{NeuroViz} through a comparative user study with 31 participants, comparing it against six established visualization tools. The study employed standardized quantitative instruments, including the System Usability Scale (SUS), NASA Task Load Index (NASA-TLX), and perceived trust ratings, alongside qualitative open-ended feedback analyzed through Reflexive Thematic Analysis. These measures assess usability, cognitive workload, trust, transparency, and user preference across all seven tools. The contributions of this work are: (1) a web-based visualization system that renders the complete feedforward training cycle with synchronized, step-aligned animations at the neuron level; (2) a comparative user study evaluating \emph{NeuroViz} against six established tools across usability, workload, trust, and interpretability; and (3) empirical evidence that synchronized training visualization improves perceived transparency without increasing cognitive load.

\section{Related Work}
\label{sec:related}

Interactive visualization has played a central role in making deep neural networks more accessible for education, model inspection, and debugging by providing human-interpretable views into otherwise opaque computation~\cite{hohman2019survey,rauber2017hidden,kahng2017activis,wongsuphasawat2018tfgraph}. In parallel, interpretability research has emphasized the need for tools that help practitioners build reliable mental models of what networks learn and how they behave~\cite{lipton2018mythos,doshi2017towards,olah2018building,olah2017featureviz}. This section describes six state-of-the-art visualization tools directly compared in our user study (Section~\ref{sec:er}) and summarizes their core features and limitations with respect to visualizing training dynamics.

\emph{Netron}~\cite{netron}: An open-source visualization tool supporting different model formats, including ONNX\footnote{\url{https://onnx.ai}}, TensorFlow\footnote{\url{https://www.tensorflow.org}}, PyTorch\footnote{\url{https://pytorch.org}}, and Core ML\footnote{\url{https://developer.apple.com/documentation/coreml}} . It renders architectures as directed graphs for inspecting layers, parameters, and connections. Netron excels at static model inspection and structural understanding, similar to other graph-based inspection approaches~\cite{wongsuphasawat2018tfgraph}, but it offers no training visualization, backpropagation animation, or real-time interaction with learning processes.

\emph{CNN Explainer}~\cite{cnnexplainer}: An educational tool designed to teach convolutional neural networks \cite{lecun1989backpropagation} to non-experts. It combines a high-level overview with interactive, on-demand views of convolutions, pooling, activations, and feature maps, with smooth transitions across levels. It is effective for CNN-specific components but is limited to pre-trained inference on fixed architectures and does not support general feedforward training or backpropagation visualization.

\emph{GAN Lab}~\cite{ganlab}: It is designed to explore Generative Adversarial Networks \cite{goodfellow2014generative}. It visualizes training through dynamic views of generator/discriminator distributions, decision boundaries (as heatmaps). Illustrates real-time interactivity for generative models, consistent with research directions in GAN training visualization~\cite{wang2018ganviz}, but is specialized for GANs and does not generalize to standard feedforward networks or backpropagation in classification/regression tasks.

\emph{TensorFlow Playground}~\cite{tfplayground}: A browser-based platform for experimenting with small feedforward networks on toy datasets. It provides real-time visualization of activations, decision boundaries, loss curves, and hyperparameter effects during training. Highly interactive and accessible, it illustrates basic concepts but does not show detailed backward propagation flows, synchronized gradient movement, intra-epoch weight update timing, or per-neuron equations.

 \emph{LLM Visualization}~\cite{bycroft_llm}: An interactive 3D walkthrough of transformer based large language models (using a small GPT-style architecture). It exposes tokenization, embeddings, self-attention, and layer-wise transformations during inference. It aids understanding of transformer mechanics \cite{vaswani2017attention} but focuses on forward-pass inference rather than training dynamics, backpropagation, or interactive weight updates.

\emph{Transformer Explainer}~\cite{transformerexplainer}: An interactive, step-by-step visualization designed to explain transformer models \cite{vaswani2017attention} at the level of individual tokens. It enables users to explore embeddings, residual connections, and layer-wise transformations through coordinated visual views. The tool is highly effective for building attention mechanisms during inference, particularly for educational purposes. However, similar to other transformer-focused explainers, it is limited to forward-pass inference on pre-defined models and does not visualize training dynamics, backpropagation, gradient flow, or weight update timing. 

\begin{table*}[t]

\centering
\caption{Comparison of existing neural network and LLM visualization tools.
\checkmark\ indicates strong support; \textbullet\ indicates partial/limited support; \texttimes\ indicates no support.}
\label{tab:tool-comparison}
\scriptsize
\setlength{\tabcolsep}{3pt}
\renewcommand{\arraystretch}{1.15}
\makebox[\linewidth][c]{%
\begin{tabular}{l c c c c c p{2.6cm}}
\toprule
Tool & Real-Time Training & \makecell{Backpropagation\\Weight Updates}  & Custom Data & Live Equations & Model Types \\

\midrule
TensorFlow Playground~\cite{tfplayground} & \checkmark & \texttimes  & \texttimes & \texttimes & Feedforward only \\
Netron~\cite{netron} & \texttimes & \texttimes & \texttimes & \texttimes & Static architecture inspection \\
CNN Explainer~\cite{cnnexplainer} & \textbullet & \texttimes  & \texttimes & \textbullet & CNNs only \\
GAN Lab~\cite{ganlab} & \checkmark & \textbullet  & \texttimes & \texttimes & GANs only \\
LLM Visualization~\cite{bycroft_llm} & \textbullet & \texttimes & \texttimes & \texttimes & Transformer-based LLMs \\
Transformer Explainer~\cite{transformerexplainer} & \texttimes & \texttimes  & \texttimes & \texttimes & Transformer inference only \\
\bottomrule
\end{tabular}}
\end{table*}

Table~\ref{tab:tool-comparison} compares these tools with \emph{NeuroViz} across key features. While several tools provide strong real-time interactivity or architectural insight, they do not offer synchronized views of learning, specifically, the coupled flow of forward activations, backward error gradients, and the exact timing of weight updates within training iterations. These visualization systems have explored internal activations and learned features~\cite{yosinski2015deepviz,hohman2020summit} and gradient-based training behavior in sequence models~\cite{cashman2018rnnbow,strobelt2018lstmvis}, yet, step-aligned depictions of backpropagation and weight update timing remain uncommon in interactive feedforward training tools. This gap motivates the design of NeuroViz, which we describe in the following section.
\medskip

\section{System Design}
\label{sec:design}

\begin{figure*}[t]
    \centering
    \includegraphics[
        width=0.6\textwidth,
        clip
    ]{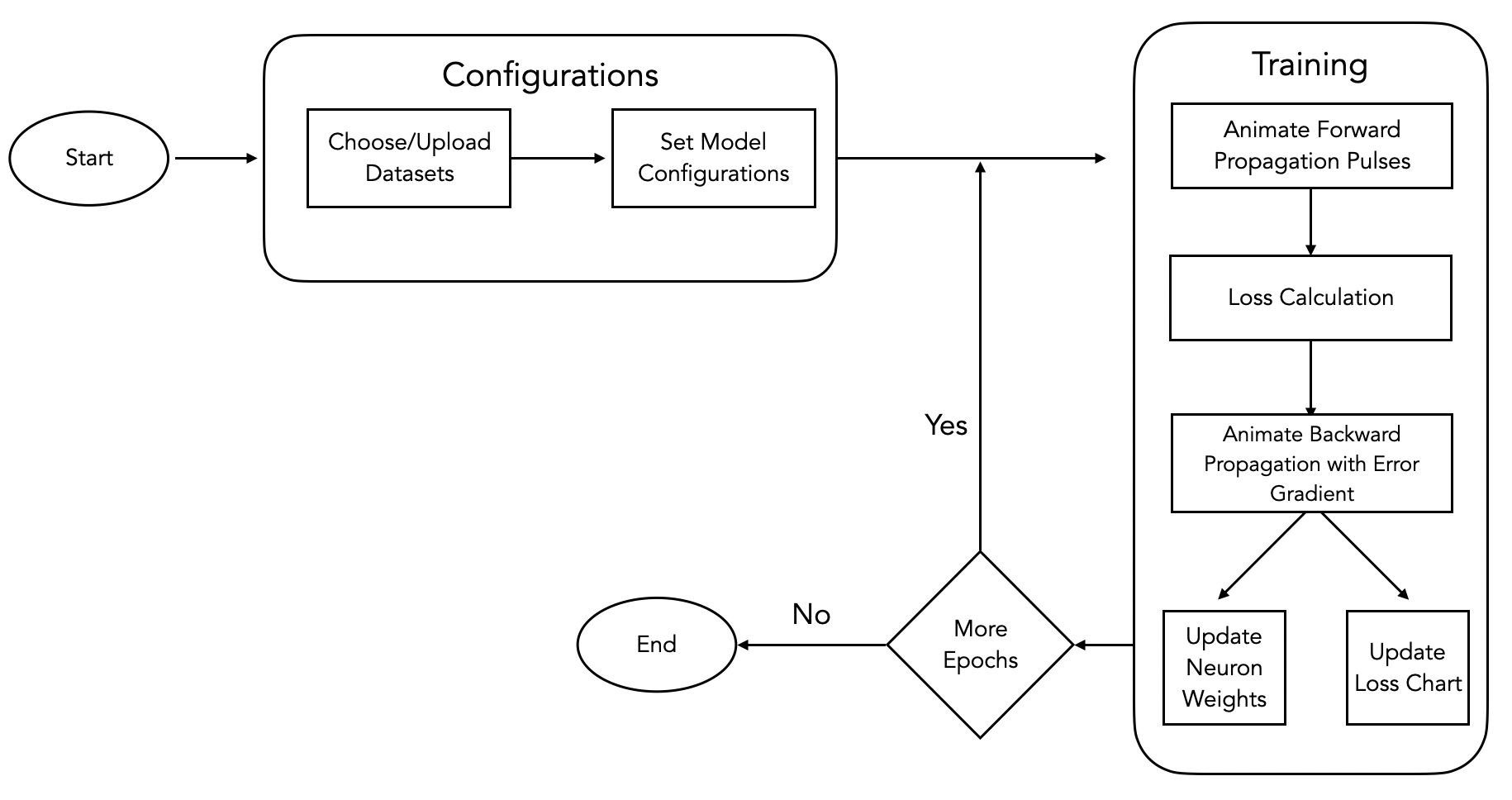}
    \caption{Process flow of \emph{NeuroViz} visualization tool}
    \label{fig:NeuroViz_FlowChart}
\end{figure*}

\emph{NeuroViz} is a browser-based interactive visualization tool implemented with React\footnote{\url{https://react.dev/}} for the user interface and TensorFlow.js \footnote{\url{https://www.tensorflow.org/js}} for neural network computation and training. The system makes the training process of feedforward neural networks fully observable and operable in real time, supporting both classification and regression tasks. The interface is organized into four primary regions, including dataset selection, model configuration, network visualization, and training progress monitoring. Figure~\ref{fig:NeuroViz_FlowChart} provides an overview of the workflow from dataset selection to training and interaction.

\subsection{Dataset Selection and Configuration}

The workflow begins with a tabular dataset selection. Users can choose between built-in sample datasets or upload their own. The built-in options include the Iris dataset (a classification task with 4 features and 3 species classes) and a Diabetes dataset (a regression task with 6 normalized features). Upon selection, the interface displays a summary of the chosen dataset, including the number of samples, feature names, target variable, and task type. For uploaded datasets, \emph{NeuroViz} automatically infers feature columns and task type from the data.

Once a dataset is selected by the user, they configure the network through a dedicated control panel. The configurable parameters include the activation function (e.g., sigmoid, ReLU), the problem type (classification or regression), the learning rate, and the number of training epochs. Users can configure the multilayer perceptron architecture by adding or removing hidden layers through an ``Add Layer'' button and adjusting the number of neurons in each layer with increment and decrement controls. Our interface also provides a set of test input fields corresponding to the dataset features, along with a ``Predict'' button that allows users to run inference on custom inputs at any point.
 
\subsection{Training Visualization}

When the user initiates training by pressing ``Play,'' the framework begins epoch-by-epoch execution with animated visualization of the complete training cycle. The network is rendered as a directed graph in which neurons are displayed as labeled nodes (input nodes labeled with feature names, hidden nodes as $h_1, h_2, \ldots$, and output nodes as $o_1, o_2, \ldots$), and connections are rendered as directed edges with numeric weight labels.

\begin{figure*}[t]
    \centering

    \begin{subfigure}[t]{0.48\textwidth}
        \centering
        \includegraphics[width=\linewidth]{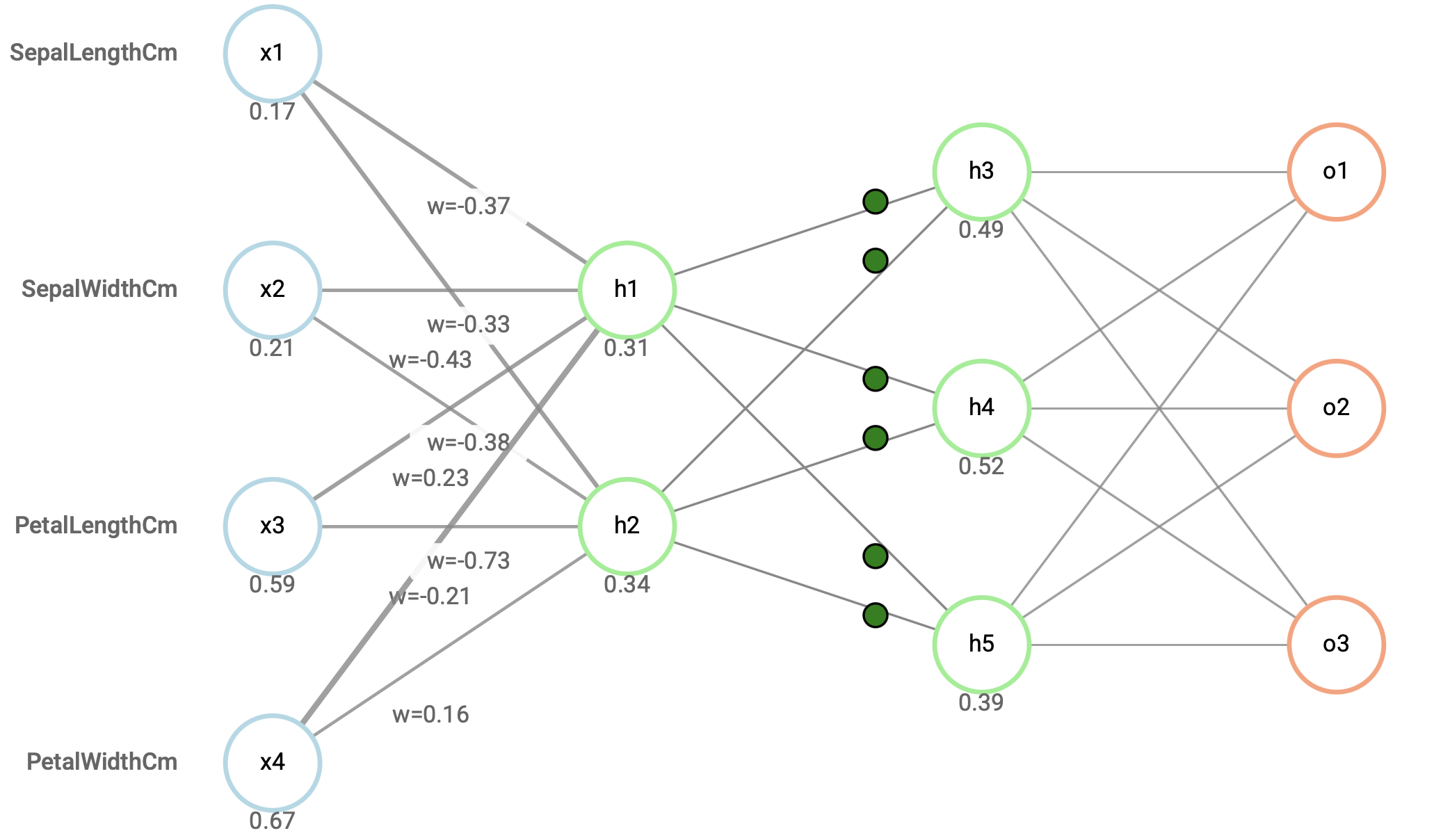}
        \caption{Forward propagation}
        \label{fig:forward}
    \end{subfigure}
    \hfill
    \begin{subfigure}[t]{0.48\textwidth}
        \centering
        \includegraphics[width=\linewidth]{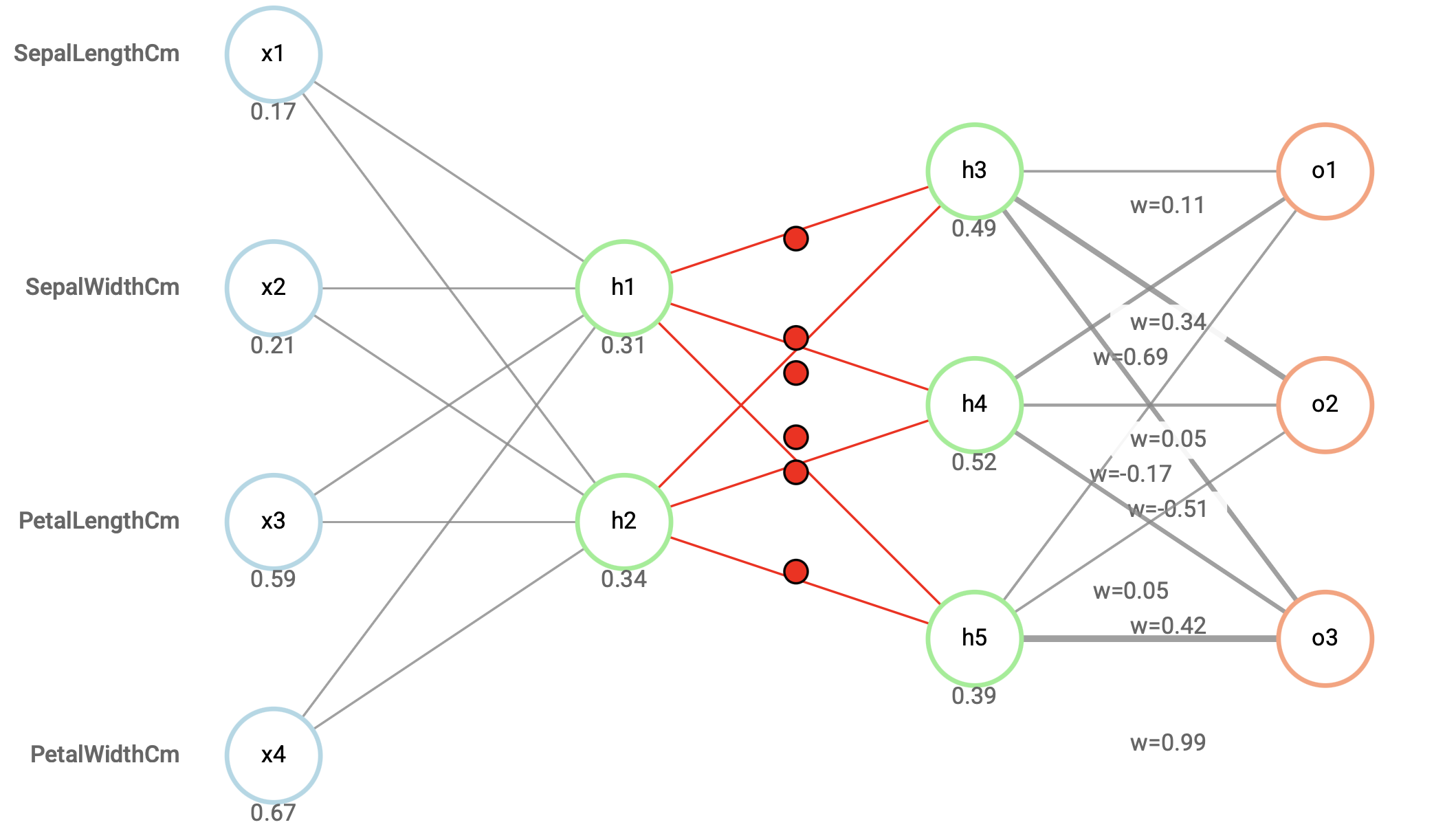}
        \caption{Backward propagation}
        \label{fig:backward}
    \end{subfigure}

    \medskip

    \begin{subfigure}[t]{0.48\textwidth}
        \centering
        \includegraphics[width=\linewidth]{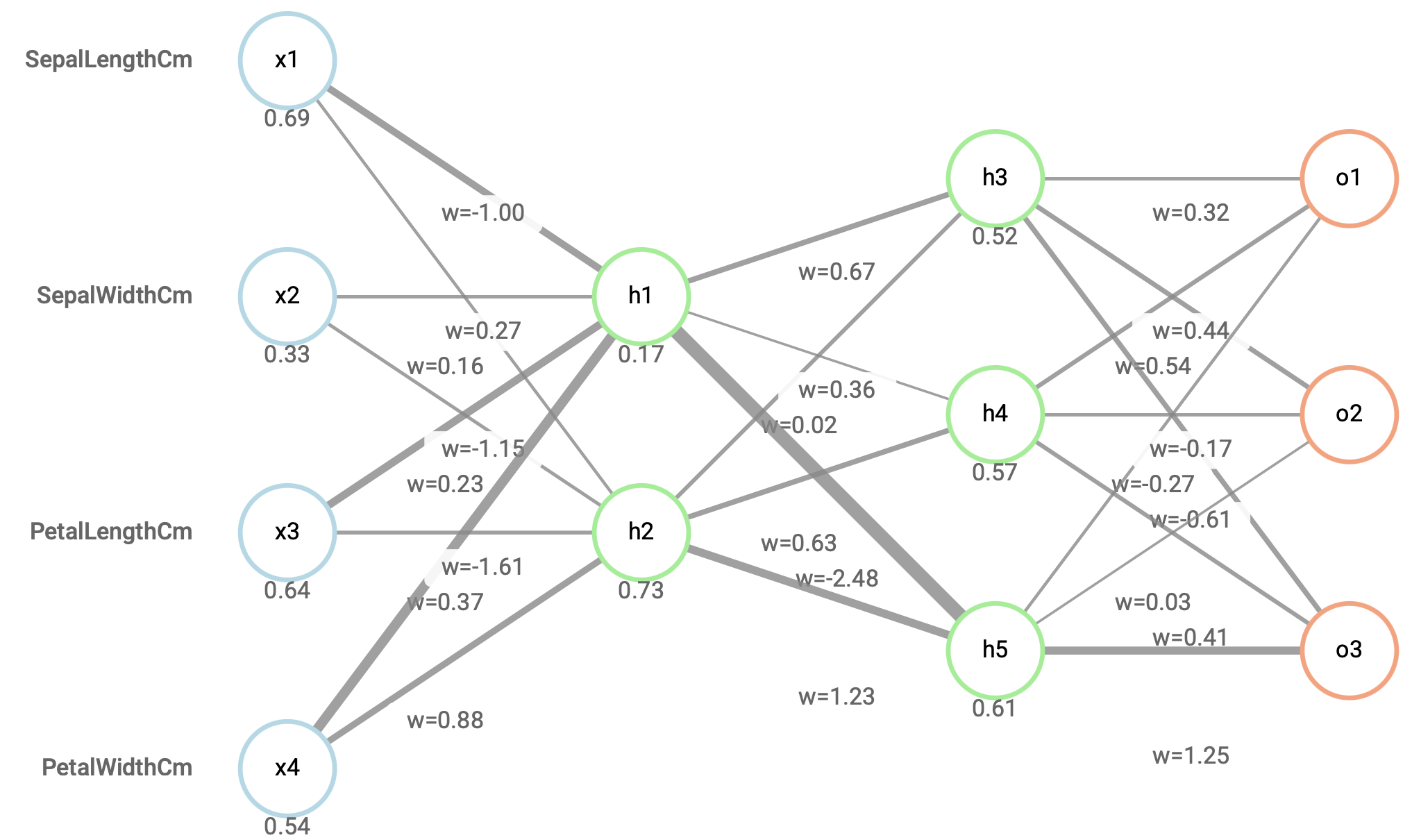}
        \caption{Weight evolution (edge thickness reflects magnitude)}
        \label{fig:weight}
    \end{subfigure}
    \hfill
    \begin{subfigure}[t]{0.48\textwidth}
        \centering
        \includegraphics[width=\linewidth]{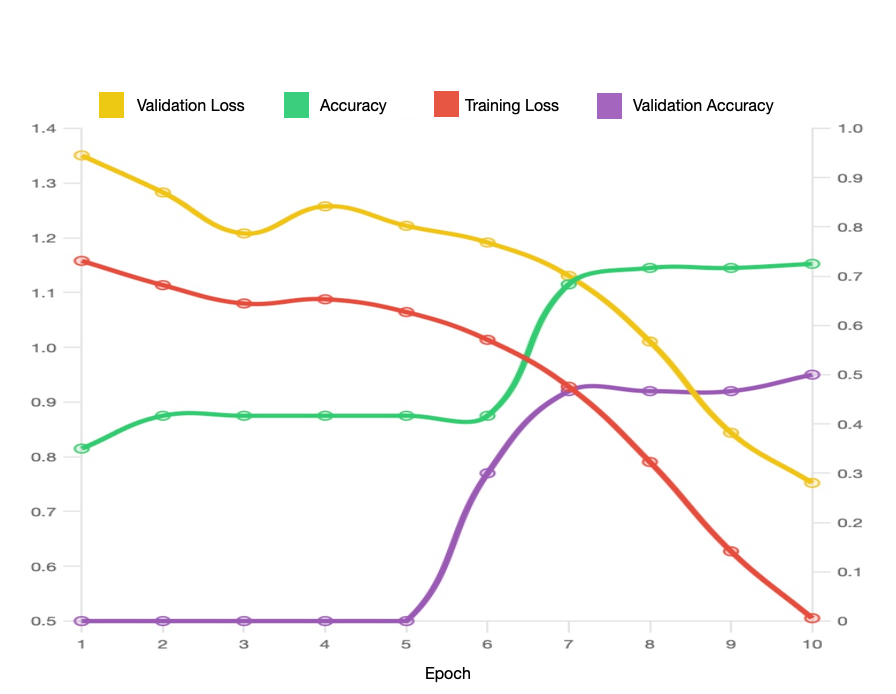}
        \caption{Training progress (loss and accuracy curves)}
        \label{fig:loss}
    \end{subfigure}

    \caption{Visualization examples from \emph{NeuroViz} on the Iris dataset.}
    \Description{Four-panel visualization showing forward propagation, backward propagation, weight evolution, and training progress.}
    \label{fig:visual-examples}
\end{figure*}

During each epoch, three phases are animated in sequence. First, forward propagation is shown through green pulses that travel along connections from the input layer toward the output layer, with each neuron displaying its current activation value (Figure~\ref{fig:forward}). Second, after loss computation, backward propagation is visualized through red pulses that travel from the output layer back toward the input layer, representing the flow of error gradients (Figure~\ref{fig:backward}). Third, weights are updated only after the corresponding error signal arrives at each neuron, allowing users to observe the distinction between pre-update and post-update parameter states within a single epoch.

Edge thickness is scaled by current weight magnitude when the ``Weight-based'' line thickness mode is selected, providing immediate visual feedback on which connections carry the most influence. 

\subsection{Monitoring and Interaction During Training}

Training can be paused, resumed, or stopped at any point using the Play, Pause, and Stop controls. A ``Network Information'' panel on the left side of the interface provides a summary of the current state, including dataset statistics, architecture details, training status, hyperparameters, and model statistics. These values update in real time as training progresses.

The ``Training Progress'' chart (Figure~\ref{fig:loss}) on the right bottom side of Figure ~\ref{fig:visual-examples}, plots loss and accuracy (or validation loss and validation accuracy, when applicable) as curves over epochs, enabling users to track convergence behavior across the full training episode.

Below the network graph, dynamically updating per-neuron equations display the learned function at each output neuron (e.g., $o_1 = \text{softmax}(0.45 \cdot h_1 + 0.28 \cdot h_2 + 0.16 + b)$), making the relationship between weights, activations, and outputs explicit. Hovering over any neuron in the network displays a tooltip showing the full weighted sum equation ($z = w_1 \cdot x_1 + w_2 \cdot x_2 + \ldots + b$) for that neuron, allowing users to inspect individual computations without navigating away from the main view.

\section{Experimental Evaluation}
\label{sec:methodology}

To evaluate our proposed approach, we conducted a comparative user study focused on usability, interpretability, and the extent to which interactive visualization supports intuition about backpropagation and weight updates.

\subsection{Objectives}
Our user study was designed to address the following objectives:
\begin{enumerate}
    \item Assess the usability of \emph{NeuroViz} relative to six established neural network visualization tools using standardized instruments.
    \item Evaluate whether the synchronized, step-aligned visualization  of forward propagation, backpropagation, and weight updates reduces perceived cognitive workload compared to existing approaches.
    \item Measure the degree to which each visualization tool fosters perceived trust and transparency in neural network training.
    \item Collect and analyze qualitative feedback on the strengths, limitations, and  design opportunities across all tools, with particular attention to interactivity, weight-update visibility, and backpropagation intuition.
\end{enumerate}

\subsection{Participants}

We recruited 31 graduate students majoring in computer science and data science. All participants had prior exposure to neural networks, learned them in their courses, and were familiar with core concepts such as forward propagation, backpropagation, activation functions, and optimization. Before starting the study, participants reported their (i) familiarity with machine learning and neural network concepts and (ii) whether they had previously used neural network visualization tools, allowing us to contextualize responses across differing levels of experience. Participation was voluntary.

Approximately 51\% of respondents identified as having intermediate experience with machine learning models, while 39\% reported advanced experience and 10\% reported basic familiarity. In addition, 61\% of participants indicated prior experience using neural network visualization tools, while 39\% reported no prior exposure.

\subsection{Apparatus}

Study sessions were conducted remotely over a video meeting software. Each participant used their own computer with a web browser (Chrome, Firefox, or Safari) to access the seven visualization tools. \emph{NeuroViz} was hosted as a web application and accessed through a shared URL. The six comparison tools, TensorFlow Playground~\cite{tfplayground}, Netron~\cite{netron}, CNN Explainer~\cite{cnnexplainer}, GAN Lab~\cite{ganlab}, LLM Visualization~\cite{bycroft_llm}, and Transformer Explainer~\cite{transformerexplainer}, were accessed through their respective web interfaces. Post-session questionnaires were administered through Google Forms. No specialized hardware or software installation was required beyond a standard web browser.

\subsection{Procedure}

We employed a within-subject comparative design ~\cite{lazar2017research}, a common approach in usability evaluation to reduce inter-participant variance and enable direct comparison across systems. Each participant interacted with all seven visualization tools (\emph{NeuroViz} and six comparison tools) in a single session. To mitigate ordering and learning effects, the order of tools was randomized across participants.

During each session, participants explored each tool for approximately 5--7 minutes. Participants were not constrained to a fixed sequence of tasks; instead, they were encouraged to interact with each tool in whatever way they believed would best enhance their understanding of the model and its training behavior. This open-ended interaction was intentional, because our goal was to evaluate how well each tool supports natural exploration and intuition-building rather than performance on narrowly defined tasks. After using all seven tools, participants completed a structured questionnaire covering both quantitative and qualitative measures.

\subsection{User Study Metrics}

The questionnaire combined standardized quantitative instruments with qualitative open-ended items.

\subsubsection{Quantitative Measures}

\paragraph{System Usability Scale (SUS).} For each tool, participants completed the System Usability Scale (SUS)~\cite{brooke1996sus}, a 10-item instrument yielding a composite score from 0 to 100. SUS is widely used for standardized comparison of perceived usability, with scores above 68 generally considered above average and scores above 80.3 classified as ``excellent''~\cite{bangor2009determining}.

\paragraph{NASA-TLX (Cognitive Workload).} Cognitive workload was assessed using an adapted, unweighted NASA Task Load Index (NASA-TLX)~\cite{hart1988development}, computing the mean across four subscales, mental demand, time pressure, effort, and frustration. Each subscale was rated on a 1--5 scale, with lower scores indicating less demanding interactions. Participants also rated perceived task accomplishment on the same scale.

\paragraph{Perceived Trust and Transparency.} Trust was measured using five items adapted from established trust in automation scales~\cite{jian2000foundations, lee2004trust}, rated on a 1--5 scale. The items captured the degree to which participants felt they could rely on the tool's outputs, trust it for decision support, found its behavior predictable, felt confident in the accuracy of its results, and perceived its behavior as transparent and understandable. The composite trust score was computed as the mean across all five items. Participants also rated additional statements on a 1--5 Likert scale~\cite{likert1932technique}, capturing ease of use, clarity of explanation, and improvement of understanding.

\paragraph{Open-ended Feedback.} At the end of the survey, participants responded to open-ended questions, including which tool they found most and least effective (with reasons), which tool they would regularly use, and what changes they would suggest. Participants also commented on whether the tools altered their perception of neural network transparency. These responses were later analyzed using Reflexive Thematic Analysis~\cite{braun2006using} to identify recurring themes.

\subsection{Aggregated Analysis}
We computed descriptive statistics for SUS, NASA-TLX, and Trust scores (mean, standard deviation, and response distributions) across all seven tools. Given the ordinal, within-subjects nature of the data, we applied the Friedman test~\cite{friedman1937use} as the omnibus test for each metric. Where the Friedman test indicated a significant effect, we conducted pairwise Wilcoxon Signed-Rank tests~\cite{wilcoxon1945individual} comparing \emph{NeuroViz} against each of the six other tools, with Bonferroni correction applied across six comparisons (adjusted $\alpha = .0083$). Effect sizes are reported as rank-biserial correlations ($r$), where 
$r > .50$ indicates a large effect~\cite{cohen1988statistical}. 

Open-ended responses were qualitatively analyzed using Reflexive Thematic 
Analysis~\cite{braun2006using} to identify recurring themes around 
interactivity, weight-update visibility, backpropagation intuition, 
and overall tool preference. The thematic findings are discussed in 
detail in Section~\ref{sec:discussion}.

\section{Evaluation Results} \label{sec:er}

\subsection{Usability Experiments}
\label{subsec:sus}

Table~\ref{tab:descriptive} summarises the mean SUS, NASA-TLX workload, and Trust scores across all seven tools. \emph{NeuroViz} achieved the highest mean SUS score ($M = 80.97$, $SD = 12.99$), placing it in the \textit{Excellent} band per Bangor et al.~\cite{bangor2009determining}. TensorFlow Playground was the closest comparison ($M = 77.82$, $SD = 13.53$), while all remaining tools fell in the \textit{Marginal} to \textit{Poor} range. Netron scored lowest ($M = 59.76$, $SD = 23.11$), consistent with its role as a static architecture viewer with no interactive training capability.

\begin{table}[!htb]
\centering
\caption{Mean ($M$) and standard deviation ($SD$) for SUS (0--100), NASA-TLX workload (1--5, lower is better), and Trust (1--5) across all tools ($N = 31$).}
\label{tab:descriptive}
\small
\setlength{\tabcolsep}{5pt}
\renewcommand{\arraystretch}{1.15}
\begin{tabular}{lcccccc}
\toprule
\multirow{2}{*}{\textbf{Tool}} &
\multicolumn{2}{c}{\textbf{SUS}} &
\multicolumn{2}{c}{\textbf{NASA-TLX}} &
\multicolumn{2}{c}{\textbf{Trust}} \\
\cmidrule(lr){2-3}\cmidrule(lr){4-5}\cmidrule(lr){6-7}
 & $M$ & $SD$ & $M$ & $SD$ & $M$ & $SD$ \\
\midrule
Netron                 & 59.76 & 23.11 & 2.38 & 0.86 & 3.63 & 0.76 \\
TensorFlow Playground  & 77.82 & 13.53 & 1.75 & 0.79 & 3.93 & 0.58 \\
\textbf{NeuroViz}      & \textbf{80.97} & 12.99 & \textbf{1.68} & 0.65 & \textbf{4.20} & 0.65 \\
CNN Explainer          & 65.48 & 19.13 & 2.27 & 0.94 & 3.94 & 0.81 \\
GAN Lab                & 64.92 & 17.92 & 2.19 & 0.81 & 3.82 & 0.78 \\
LLM Visualisation      & 67.18 &  8.82 & 2.11 & 0.41 & 3.85 & 0.36 \\
Transformer Explainer  & 66.85 &  7.98 & 2.15 & 0.47 & 3.89 & 0.37 \\
\bottomrule
\end{tabular}
\end{table}

Given the ordinal, within-subjects nature of the data, we applied the Friedman test~\cite{friedman1937use}, which revealed a significant effect of tool on SUS scores ($\chi^2(6) = 33.996$, $p < .001$, Kendall's $W = 0.183$). We then conducted pairwise Wilcoxon Signed-Rank post-hoc tests~\cite{wilcoxon1945individual} comparing \emph{NeuroViz} against each tool, with Bonferroni correction applied across six comparisons (adjusted $\alpha = .0083$). Effect sizes are reported as rank-biserial correlations ($r$), where $r > .50$ indicates a large effect~\cite{cohen1988statistical}.

As shown in Table~\ref{tab:posthoc_sus}, \emph{NeuroViz} significantly outperformed four tools on usability at the corrected threshold, i.e., Netron ($p_{\text{adj}} = .0023$), GAN Lab ($p_{\text{adj}} = .0005$), LLM Visualisation ($p_{\text{adj}} = .0024$), and Transformer Explainer ($p_{\text{adj}} = .0017$), with large effect sizes throughout ($r = 0.728$ to $0.852$). The comparison against CNN Explainer narrowly missed the corrected threshold ($p_{\text{adj}} = .0098$) but still showed a large effect ($r = 0.658$), suggesting a practically meaningful difference. No significant difference was found against TensorFlow Playground ($p_{\text{adj}} = 1.000$), indicating these two tools occupy a comparable usability tier.

\begin{table}[!htb]
\centering
\caption{Wilcoxon Signed-Rank post-hoc tests of \emph{NeuroViz} vs.\ each tool on SUS. $r$ = rank-biserial correlation.}
\label{tab:posthoc_sus}
\small
\setlength{\tabcolsep}{5pt}
\renewcommand{\arraystretch}{1.15}
\begin{tabular}{lcccc}
\toprule
\textbf{Comparison} & $W$ & $p$ & $p_{\text{adj}}$ & $r$ \\
\midrule
vs.\ Netron                  & 47  & .0004 & .0023 & 0.768 \\
vs.\ TensorFlow Playground   & 158 & .455  & 1.000 & 0.164 \\
vs.\ CNN Explainer           & 80  & .0016 & .0098 & 0.658 \\
vs.\ GAN Lab                 & 30  & .0001 & .0005 & 0.852 \\
vs.\ LLM Visualisation       & 68  & .0004 & .0024 & 0.728 \\
vs.\ Transformer Explainer   & 63  & .0003 & .0017 & 0.746 \\
\bottomrule
\multicolumn{5}{l}{\footnotesize Significant comparisons satisfy $p_{\text{adj}} < .0083$ (Bonferroni, 6 comparisons); ns otherwise.}
\end{tabular}
\end{table}

\begin{figure*}[t!]
    \centering
    \includegraphics[width=0.32\textwidth]{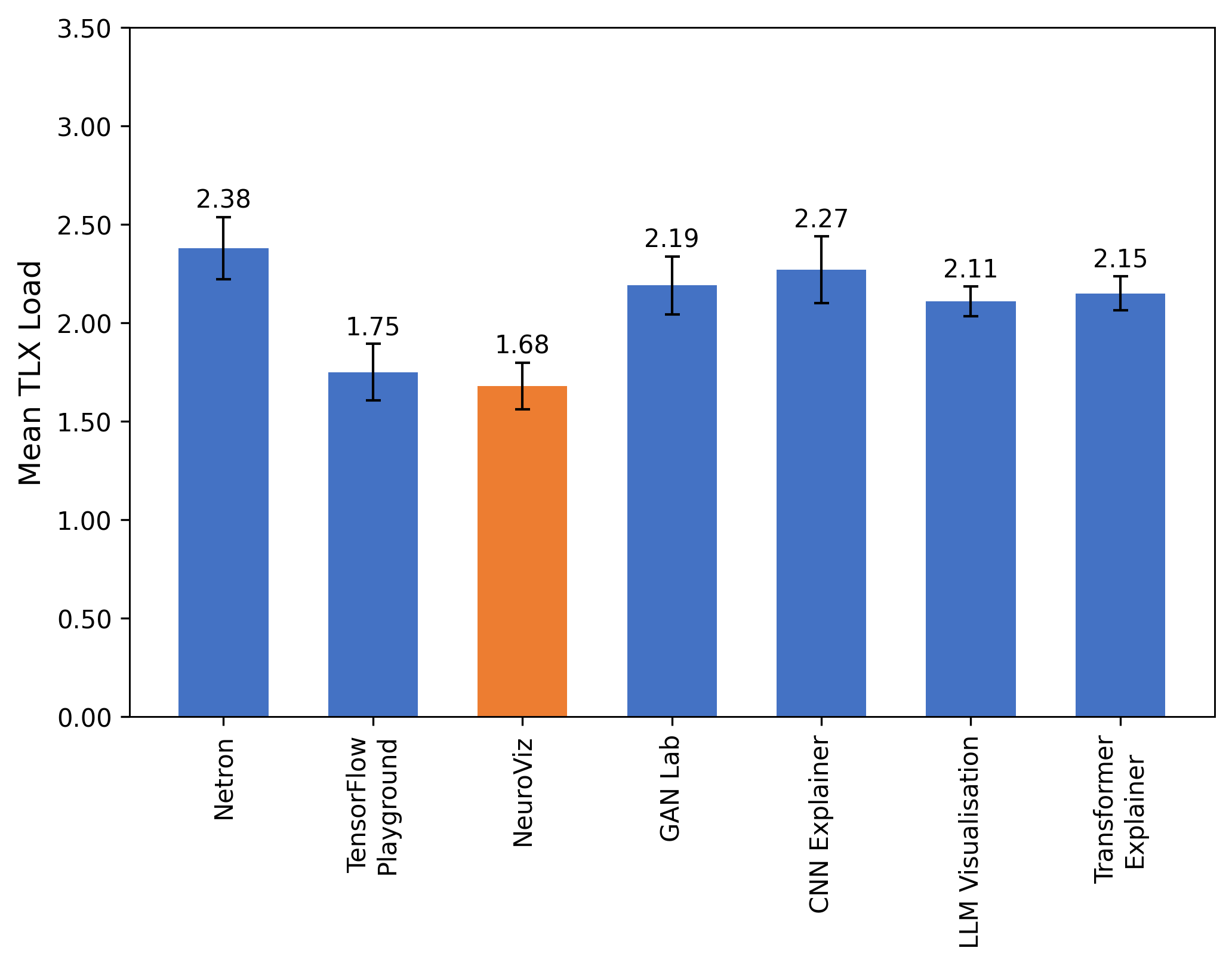}
    \hfill
    \includegraphics[width=0.32\textwidth]{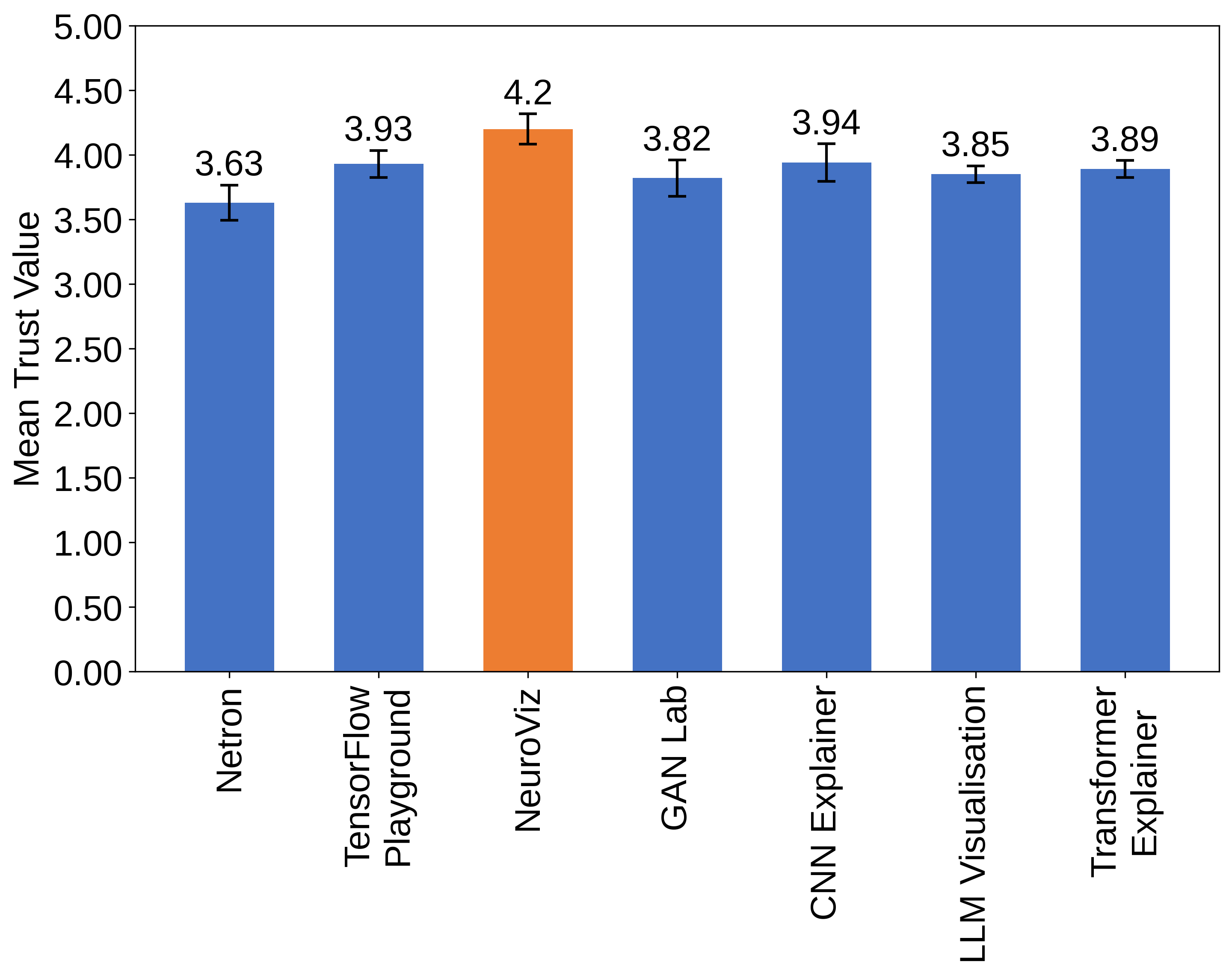}
    \hfill
    \includegraphics[width=0.32\textwidth]{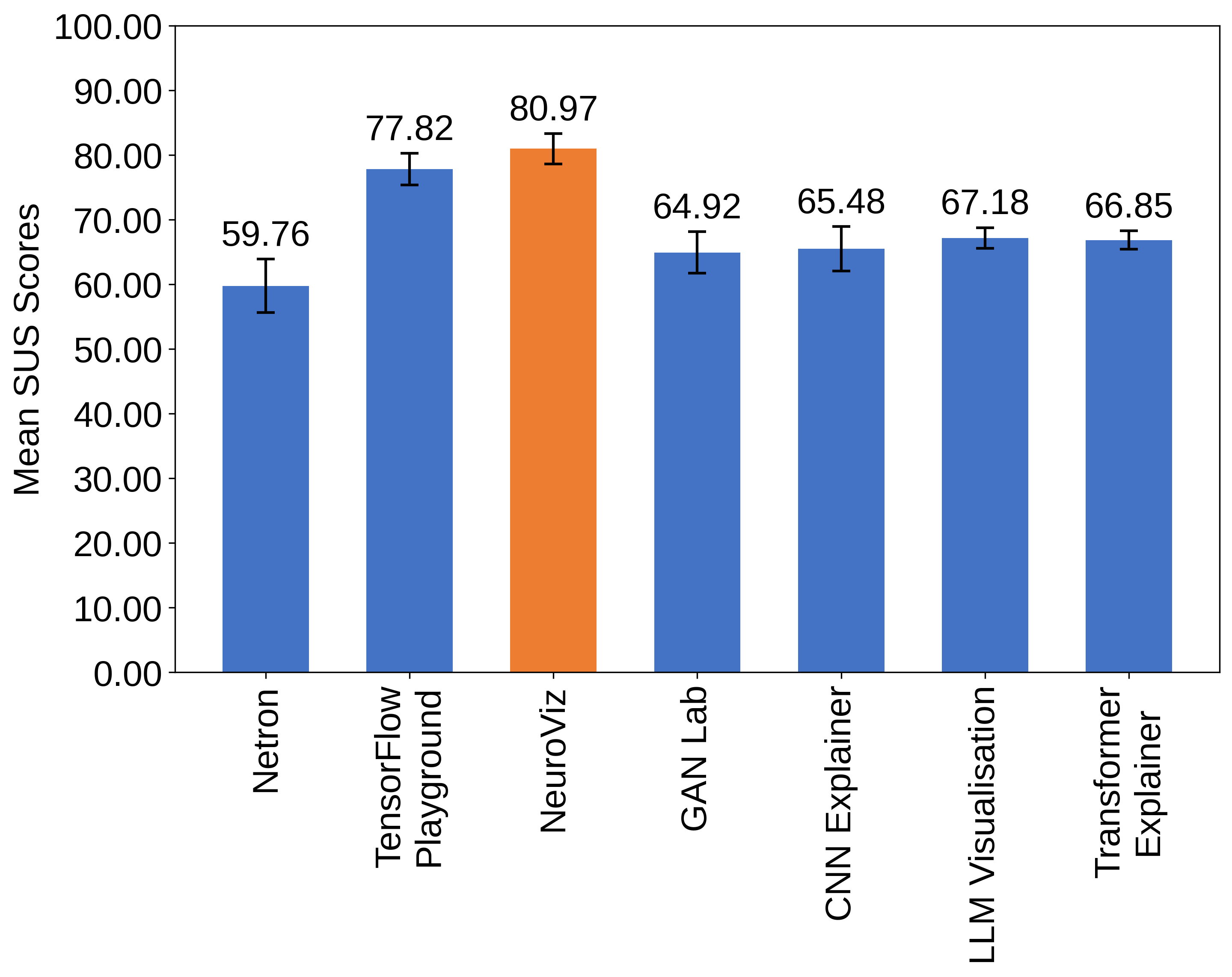}
    \caption{Comparison of cognitive load (left), perceived trust (center), and usability (right) across tools.}
    \Description{Three bar charts comparing metrics across tools.}
    \label{fig:composite-eval}
\end{figure*}

\subsection{Cognitive Workload (NASA-TLX)}
\label{subsec:tlx}

Cognitive workload was assessed using an adapted (unweighted) NASA-TLX~\cite{hart1988development}, computing the mean across four sub-scales: mental demand, time pressure, effort, and frustration. Lower scores indicate less demanding interactions. \emph{NeuroViz} produced the lowest workload of any tool ($M = 1.68$, $SD = 0.65$), alongside the highest perceived task accomplishment ($M = 4.35$, $SD = 0.70$). This combination suggests that NeuroViz's structured visualization of the training cycle reduces cognitive effort without sacrificing depth~\cite{sweller1994cognitive}.

The Friedman test confirmed a significant effect across tools ($\chi^2(6) = 28.123$, $p < .001$, Kendall's $W = 0.151$). As shown in Table~\ref{tab:posthoc_tlx}, while \emph{NeuroViz} consistently showed lower workload than all tools except TensorFlow Playground, none of these differences survived the strict Bonferroni correction ($\alpha = .0083$). However, the large effect sizes observed ($r = 0.568$ to $0.672$) indicate practically meaningful differences that a larger sample may confirm statistically.

\begin{table}[!htb]
\centering
\caption{Wilcoxon Signed-Rank post-hoc tests: \emph{NeuroViz} vs.\ each tool on NASA-TLX workload (adapted, unweighted). Lower workload is better.}
\label{tab:posthoc_tlx}
\small
\setlength{\tabcolsep}{5pt}
\renewcommand{\arraystretch}{1.15}
\begin{tabular}{lcccc}
\toprule
\textbf{Comparison} & $W$ & $p$ & $p_{\text{adj}}$ & $r$ \\
\midrule
vs.\ Netron                  & 66  & .0018 & .0107 & 0.672 \\
vs.\ TensorFlow Playground   & 174 & .725  & 1.000 & 0.077 \\
vs.\ CNN Explainer           & 43  & .0066 & .0396 & 0.660 \\
vs.\ GAN Lab                 & 55  & .0037 & .0223 & 0.662 \\
vs.\ LLM Visualisation       & 97  & .0050 & .0302 & 0.583 \\
vs.\ Transformer Explainer   & 94  & .0072 & .0434 & 0.568 \\
\bottomrule
\multicolumn{5}{l}{\footnotesize Significant comparisons satisfy $p_{\text{adj}} < .0083$; none meet this threshold here.}
\end{tabular}
\end{table}

\subsection{Perceived Trust}
\label{subsec:trust}

\emph{NeuroViz} achieved the highest mean trust rating of any tool ($M = 4.20$, $SD = 0.65$), with particularly strong scores on the transparency item ($M = 4.29$). The Friedman test revealed a significant overall effect ($\chi^2(6) = 13.777$, $p = .032$, Kendall's $W = 0.074$). Post-hoc testing (Table~\ref{tab:posthoc_trust}) showed a statistically significant advantage over Netron at the corrected threshold ($p_{\text{adj}} = .0048$, $r = 0.724$). Differences against the remaining tools did not meet the corrected threshold, though consistent medium-to-large effect sizes ($r = 0.421$ to $0.522$) suggest a directional advantage that a larger sample may confirm~\cite{lee2004trust}.

\begin{table}[!htb]
\centering
\caption{Wilcoxon Signed-Rank post-hoc tests: \emph{NeuroViz} vs.\ each tool on Trust (1--5).}
\label{tab:posthoc_trust}
\small
\setlength{\tabcolsep}{5pt}
\renewcommand{\arraystretch}{1.15}
\begin{tabular}{lcccc}
\toprule
\textbf{Comparison} & $W$ & $p$ & $p_{\text{adj}}$ & $r$ \\
\midrule
vs.\ Netron                  & 56  & .0008 & .0048 & 0.724 \\
vs.\ TensorFlow Playground   & 91  & .054  & .324  & 0.440 \\
vs.\ CNN Explainer           & 64  & .073  & .435  & 0.446 \\
vs.\ GAN Lab                 & 97  & .016  & .093  & 0.522 \\
vs.\ LLM Visualisation       & 122 & .039  & .233  & 0.439 \\
vs.\ Transformer Explainer   & 118 & .051  & .308  & 0.421 \\
\bottomrule
\multicolumn{5}{l}{\footnotesize Significant comparisons satisfy $p_{\text{adj}} < .0083$; ns otherwise.}
\end{tabular}
\end{table}

\subsection{Comparative Rankings}
\label{subsec:rankings}

In addition to the standardized instruments, participants ranked five of the tools (Netron, TensorFlow Playground, \emph{NeuroViz}, GAN Lab, and CNN Explainer) on four dimensions: clarity of explanations, usefulness for understanding the model, trustworthiness of explanations, and overall performance. Rankings used a 1 (best) to 5 (worst) scale. Table~\ref{tab:rankings} reports the mean rankings across all participants.

\emph{NeuroViz} achieved the highest mean ranking for usefulness ($M = 2.23$), indicating that participants found it most helpful for understanding model behavior. On clarity, \emph{NeuroViz} tied with CNN Explainer ($M = 2.47$ each), both trailing TensorFlow Playground ($M = 2.33$). TensorFlow Playground received the highest ranking for trustworthiness ($M = 2.23$). \emph{NeuroViz} ranked highest for overall performance ($M = 2.27$), with TensorFlow Playground close behind. Netron was consistently ranked lowest across all four dimensions, with mean rankings between 3.77 and 4.00.

\begin{table}[t]
\centering
\caption{Mean comparative rankings across four dimensions (1 = best, 5 = worst; $N = 30$). Lower values indicate more favorable rankings.}
\label{tab:rankings}
\small
\setlength{\tabcolsep}{5pt}
\renewcommand{\arraystretch}{1.15}
\begin{tabular}{lcccc}
\toprule
\textbf{Tool} & \textbf{Clarity} & \textbf{Usefulness} & \textbf{Trust} & \textbf{Overall} \\
\midrule
Netron                 & 4.00 & 3.87 & 3.77 & 3.93 \\
TensorFlow Playground  & \textbf{2.33} & 2.30 & \textbf{2.23} & 2.40   \\
\textbf{NeuroViz}      & 2.47 & \textbf{2.23} & 2.57 & \textbf{2.27} \\
GAN Lab                & 2.87 & 2.93 & 2.87 & 2.90 \\
CNN Explainer          & 2.47 & 2.70 & 2.50 & 2.63 \\
\bottomrule
\end{tabular}
\end{table}

\subsection{Open-Ended Feedback}
\label{subsec:qualitative}

Open-ended responses were analyzed using Reflexive Thematic Analysis~\cite{braun2006using}, identifying four recurring themes. We report the descriptive findings here; a detailed interpretation is provided in Section~\ref{sec:discussion}.

\paragraph{Most and least effective tools.} When asked to identify the most effective tool overall, TensorFlow Playground was cited most frequently (12 mentions), followed by \emph{NeuroViz} (10 mentions). Participants who favored \emph{NeuroViz} highlighted its ability to visualize backpropagation and its clarity in showing how learning unfolds across epochs. Participants who chose TensorFlow Playground emphasized its polished interface and ease of parameter manipulation. When asked about the least effective tool, Netron was cited most frequently (16 mentions), with participants noting its lack of interactivity, training visualization, and explanatory context. GAN Lab was the second most frequently cited (6 mentions), primarily due to difficulty understanding adversarial training dynamics.

\paragraph{Tool for regular use.} When asked which tool they would use regularly, \emph{NeuroViz} was the most frequently mentioned (10 mentions), followed by TensorFlow Playground (9) and CNN Explainer (8). Seven participants noted that \emph{NeuroViz}'s ability to upload custom datasets and visualize the full training cycle made it practical for ongoing use, while TensorFlow Playground was favored for its polished interface and accessibility.

\paragraph{Perception of transparency.} Of the 26 participants who provided a clear response regarding whether the tools changed their perception of neural network transparency, 19 (73\%) responded affirmatively. Participants described moving from perceiving neural networks as ``black boxes'' to understanding the step-by-step mechanics of training. Three participants (12\%) reported no change, attributing this to pre-existing knowledge, and four participants provided mixed or partial responses.

\paragraph{Recurring suggestions.} Across all tools, participants expressed a desire for embedded explanations and contextual guidance, particularly for users new to machine learning. For \emph{NeuroViz} specifically, participants suggested improving the interface layout so that the network visualization remains visible while adjusting configuration parameters, and adding tooltips or brief descriptions for each setting. Multiple participants also suggested extending \emph{NeuroViz} to support uploading pre-trained models and providing diagnostic feedback on training issues such as vanishing gradients.

\section{Discussion}
\label{sec:discussion}

\emph{NeuroViz} achieved the highest usability score (SUS = 80.97), the lowest cognitive workload (NASA-TLX $M = 1.68$), and the highest trust rating ($M = 4.20$) among all seven tools evaluated. It significantly outperformed four tools on usability at the Bonferroni-corrected threshold, with large effect sizes ($r = 0.728$ to $0.852$), and matched TensorFlow Playground on all three metrics. This outcome is notable given that TensorFlow Playground is a mature, widely adopted platform maintained by Google Brain, while \emph{NeuroViz} is a newly developed tool. The comparable performance suggests that the addition of synchronized backpropagation animation and per-neuron equations does not come at the cost of usability or increased cognitive burden.

The NASA-TLX results further support this interpretation. Although pairwise workload differences did not survive Bonferroni correction, the uniformly large effect sizes ($r = 0.568$ to $0.672$) point to practically meaningful advantages that a larger sample may confirm statistically. The combination of lowest workload and highest perceived task accomplishment ($M = 4.35$) suggests that \emph{NeuroViz}'s structured rendering of the training cycle reduces cognitive effort without sacrificing depth of information.

The trust results are consistent with this pattern. \emph{NeuroViz}'s particularly strong score on the transparency item ($M = 4.29$) indicates that exposing the step-by-step timing of signal propagation and weight updates contributes directly to user confidence in the tool's representations. This aligns with prior work arguing that mechanistic transparency is a prerequisite for trust in AI systems~\cite{doshi2017towards, lipton2018mythos}.

In the comparative rankings, \emph{NeuroViz} was rated the most useful tool for understanding model behavior ($M = 2.23$) and was the tool most frequently selected for regular use (10 of 30 participants). While TensorFlow Playground led on trustworthiness rankings, \emph{NeuroViz}'s advantage on usefulness suggests that participants valued the deeper visibility into training dynamics that \emph{NeuroViz} provides, even when they found TensorFlow Playground's interface more polished overall. Qualitative feedback identified interactivity, mechanistic transparency, and real-time weight-update visibility as the primary factors driving these outcomes. These results support the conclusion that synchronized, step-aligned visualization of the neural network training cycle can improve perceived interpretability and transparency without sacrificing usability.

Open-ended responses analyzed through Reflexive Thematic Analysis~\cite{braun2006using} yielded four recurring themes that provide further context for these quantitative findings. First, tools that allowed real-time parameter manipulation were consistently rated clearer and more insightful. \emph{NeuroViz} and TensorFlow Playground were most frequently cited as effective (8 and 12 mentions, respectively), with participants noting that watching weights update in response to gradients made backpropagation feel more concrete. As one participant noted, \emph{NeuroViz} provided ``clarity in showing what was going on during neural network training for forward and back propagation.'' Netron was identified as the least effective tool by 16 of 30 participants, with respondents describing it as offering structural detail without explanatory narrative. This finding reinforces the value of interactive, real-time visualization for building mental models of training dynamics, consistent with observations in prior visualization research~\cite{hohman2019survey}.

Second, participants expressed higher confidence in tools that exposed internal operations directly. \emph{NeuroViz}'s pulse-based animation of forward activations and backward gradients was noted as helping participants verify expected network behavior and understand the temporal sequence of learning. Several participants described transitioning from viewing neural networks as ``black boxes'' to understanding the step-by-step mechanics of training. This theme aligns with the quantitative trust results, where \emph{NeuroViz} scored highest on the transparency item ($M = 4.29$), and with transparency-based accounts of trust in interactive systems~\cite{lee2004trust}.

Third, a recurring tension emerged between approachability and depth. Participants with intermediate machine learning experience favored \emph{NeuroViz} and TensorFlow Playground for making complexity manageable. One participant observed that if 
``NeuroViz's UI was more polished with better UX practices, it would probably be at the top of my list,'' suggesting that the tool's content and visualization approach are strong but that interface refinement could further improve its standing. GAN Lab's interactivity was seen as engaging but confusing (cited as least effective by 6 participants), with generative outputs not translating into comprehension for participants unfamiliar with adversarial training. This suggests that visualization tools benefit from progressive disclosure of complexity, presenting core information by default while making deeper detail available on demand.

Fourth, across all tools, participants expressed a desire for embedded explanations and progressive disclosure of technical detail. For \emph{NeuroViz} specifically, multiple participants suggested adding tooltips or brief descriptions for each setting, inline annotations explaining what each visualization element represents,and step-by-step walkthroughs for new users. One participant noted that ``a newcomer wouldn't necessarily understand what was going on'' without additional guidance. Six participants also suggested extending \emph{NeuroViz} to support uploading pre-trained models and providing diagnostic feedback on issues such as vanishing gradients. These suggestions represent concrete design opportunities for future iterations and align with prior HCI work on layered educational interfaces~\cite{shneiderman1996eyes}.

\subsection{Limitations}
Our framework has some limitations. First, the sample consisted of 31 graduate students recruited from graduate advanced machine learning and neural network courses, which may limit the generalizability of the results to broader populations, including undergraduate students, self-taught practitioners, or domain experts outside computer science. Second, the within-subject design required participants to use seven tools in a single session, with approximately 5 to 7 minutes per tool. This limited interaction time may not have been sufficient for participants to fully explore complex tools, potentially disadvantaging tools with steeper 
learning curves. Third, \emph{NeuroViz} is currently limited to feedforward neural networks and does not support convolutional, recurrent, or transformer architectures, which restricts its applicability to a subset of modern deep learning models. Fourth, 
the comparative rankings covered only five of the seven tools, as LLM Visualization and Transformer Explainer were evaluated in a separate data collection phase and were not included in the ranking questions.

\section{Conclusion and Future Work}
\label{sec:conclusion}

We presented NeuroViz, a web-based interactive visualization tool that renders the complete training cycle of feedforward neural networks in real time, with synchronized pulse-based animations of forward activations, backward error gradients, and weight updates at the neuron level. A comparative user study with 31 participants 
evaluated NeuroViz against six established tools, where it achieved the highest usability score (SUS = 80.97, ``excellent'' range), the lowest cognitive workload, and the highest trust rating, significantly outperforming four tools with large effect sizes and matching TensorFlow Playground on all metrics. NeuroViz was rated the most useful tool for understanding model behavior and was most frequently selected for regular use, with 73\% of participants reporting increased perception of training transparency. Future work includes extending NeuroViz to support convolutional and recurrent architectures, incorporating embedded explanations and 
beginner walkthroughs as participants requested, improving the interface layout so the visualization remains visible during configuration, supporting upload of pre-trained models with diagnostic feedback on training issues, and conducting a 
longitudinal study with a broader participant pool to evaluate learning outcomes over time.

\bibliographystyle{ACM-Reference-Format}

\begin{thebibliography}{48}

%%% the Web address.

\ifx \showCODEN    \undefined \def \showCODEN     #1{\unskip}     \fi
\ifx \showDOI      \undefined \def \showDOI       #1{#1}\fi
\ifx \showISBNx    \undefined \def \showISBNx     #1{\unskip}     \fi
\ifx \showISBNxiii \undefined \def \showISBNxiii  #1{\unskip}     \fi
\ifx \showISSN     \undefined \def \showISSN      #1{\unskip}     \fi
\ifx \showLCCN     \undefined \def \showLCCN      #1{\unskip}     \fi
\ifx \shownote     \undefined \def \shownote      #1{#1}          \fi
\ifx \showarticletitle \undefined \def \showarticletitle #1{#1}   \fi
\ifx \showURL      \undefined \def \showURL       {\relax}        \fi
% invisible to TeX
\providecommand\bibfield[2]{#2}
\providecommand\bibinfo[2]{#2}
\providecommand\natexlab[1]{#1}
\providecommand\showeprint[2][]{arXiv:#2}

\bibitem[Bangor et~al\mbox{.}(2009)]%
        {bangor2009determining}
\bibfield{author}{\bibinfo{person}{Aaron Bangor}, \bibinfo{person}{Philip Kortum}, {and} \bibinfo{person}{James Miller}.} \bibinfo{year}{2009}\natexlab{}.
\newblock \showarticletitle{Determining what individual SUS scores mean: Adding an adjective rating scale}.
\newblock \bibinfo{journal}{\emph{Journal of Usability Studies}} \bibinfo{volume}{4}, \bibinfo{number}{3} (\bibinfo{year}{2009}), \bibinfo{pages}{114--123}.
\newblock


\bibitem[Bishop(2006)]%
        {bishop2006pattern}
\bibfield{author}{\bibinfo{person}{Christopher Bishop}.} \bibinfo{year}{2006}\natexlab{}.
\newblock \bibinfo{booktitle}{\emph{Pattern Recognition and Machine Learning}}.
\newblock \bibinfo{publisher}{Springer}.
\newblock


\bibitem[Braun and Clarke(2006)]%
        {braun2006using}
\bibfield{author}{\bibinfo{person}{Virginia Braun} {and} \bibinfo{person}{Victoria Clarke}.} \bibinfo{year}{2006}\natexlab{}.
\newblock \showarticletitle{Using thematic analysis in psychology}.
\newblock \bibinfo{journal}{\emph{Qualitative Research in Psychology}} \bibinfo{volume}{3}, \bibinfo{number}{2} (\bibinfo{year}{2006}), \bibinfo{pages}{77--101}.
\newblock


\bibitem[Brooke(1996)]%
        {brooke1996sus}
\bibfield{author}{\bibinfo{person}{John Brooke}.} \bibinfo{year}{1996}\natexlab{}.
\newblock \showarticletitle{SUS: A ``quick and dirty'' usability scale}. In \bibinfo{booktitle}{\emph{Usability Evaluation in Industry}}, \bibfield{editor}{\bibinfo{person}{Patrick~W. Jordan}, \bibinfo{person}{Bruce Thomas}, \bibinfo{person}{Ian~L. McClelland}, {and} \bibinfo{person}{Bernard Weerdmeester}} (Eds.). \bibinfo{publisher}{Taylor \& Francis}, \bibinfo{pages}{189--194}.
\newblock


\bibitem[Bycroft(2024)]%
        {bycroft_llm}
\bibfield{author}{\bibinfo{person}{Brendan Bycroft}.} \bibinfo{year}{2024}\natexlab{}.
\newblock \bibinfo{title}{LLM Visualization}.
\newblock \bibinfo{howpublished}{\url{https://bbycroft.net/llm}}.
\newblock
\newblock
\shownote{Interactive visualization of transformer-based large language models}.


\bibitem[Cashman et~al\mbox{.}(2018)]%
        {cashman2018rnnbow}
\bibfield{author}{\bibinfo{person}{Dylan Cashman}, \bibinfo{person}{Genevieve Patterson}, \bibinfo{person}{Abigail Mosca}, \bibinfo{person}{Nathan Watts}, \bibinfo{person}{Shannon Robinson}, {and} \bibinfo{person}{Remco Chang}.} \bibinfo{year}{2018}\natexlab{}.
\newblock \showarticletitle{RNNbow: Visualizing Learning via Backpropagation Gradients in Recurrent Neural Networks}.
\newblock \bibinfo{journal}{\emph{IEEE Computer Graphics and Applications}} (\bibinfo{year}{2018}).
\newblock
\urldef\tempurl%
\url{https://doi.org/10.1109/MCG.2018.2878902}
\showDOI{\tempurl}


\bibitem[Castelvecchi(2016)]%
        {castelvecchi2016blackbox}
\bibfield{author}{\bibinfo{person}{Davide Castelvecchi}.} \bibinfo{year}{2016}\natexlab{}.
\newblock \showarticletitle{Can we open the black box of AI?}
\newblock \bibinfo{journal}{\emph{Nature}} (\bibinfo{year}{2016}).
\newblock


\bibitem[Cohen(1988)]%
        {cohen1988statistical}
\bibfield{author}{\bibinfo{person}{Jacob Cohen}.} \bibinfo{year}{1988}\natexlab{}.
\newblock \bibinfo{booktitle}{\emph{Statistical Power Analysis for the Behavioral Sciences} (\bibinfo{edition}{2} ed.)}.
\newblock \bibinfo{publisher}{Lawrence Erlbaum Associates}.
\newblock


\bibitem[Devlin et~al\mbox{.}(2019)]%
        {devlin2019bert}
\bibfield{author}{\bibinfo{person}{Jacob Devlin} {et~al\mbox{.}}} \bibinfo{year}{2019}\natexlab{}.
\newblock \showarticletitle{BERT: Pre-training of deep bidirectional transformers for language understanding}.
\newblock \bibinfo{journal}{\emph{NAACL}} (\bibinfo{year}{2019}).
\newblock


\bibitem[Doshi-Velez and Kim(2017)]%
        {doshi2017towards}
\bibfield{author}{\bibinfo{person}{Finale Doshi-Velez} {and} \bibinfo{person}{Been Kim}.} \bibinfo{year}{2017}\natexlab{}.
\newblock \showarticletitle{Towards a rigorous science of interpretable machine learning}.
\newblock \bibinfo{journal}{\emph{arXiv preprint arXiv:1702.08608}} (\bibinfo{year}{2017}).
\newblock


\bibitem[Friedman(1937)]%
        {friedman1937use}
\bibfield{author}{\bibinfo{person}{Milton Friedman}.} \bibinfo{year}{1937}\natexlab{}.
\newblock \showarticletitle{The use of ranks to avoid the assumption of normality implicit in the analysis of variance}.
\newblock \bibinfo{journal}{\emph{J. Amer. Statist. Assoc.}} \bibinfo{volume}{32}, \bibinfo{number}{200} (\bibinfo{year}{1937}), \bibinfo{pages}{675--701}.
\newblock


\bibitem[Goodfellow et~al\mbox{.}(2016)]%
        {goodfellow2016deep}
\bibfield{author}{\bibinfo{person}{Ian Goodfellow}, \bibinfo{person}{Yoshua Bengio}, {and} \bibinfo{person}{Aaron Courville}.} \bibinfo{year}{2016}\natexlab{}.
\newblock \bibinfo{booktitle}{\emph{Deep Learning}}.
\newblock \bibinfo{publisher}{MIT Press}.
\newblock


\bibitem[Goodfellow et~al\mbox{.}(2014)]%
        {goodfellow2014generative}
\bibfield{author}{\bibinfo{person}{Ian~J Goodfellow}, \bibinfo{person}{Jean Pouget-Abadie}, \bibinfo{person}{Mehdi Mirza}, \bibinfo{person}{Bing Xu}, \bibinfo{person}{David Warde-Farley}, \bibinfo{person}{Sherjil Ozair}, \bibinfo{person}{Aaron Courville}, {and} \bibinfo{person}{Yoshua Bengio}.} \bibinfo{year}{2014}\natexlab{}.
\newblock \showarticletitle{Generative adversarial nets}.
\newblock \bibinfo{journal}{\emph{Advances in neural information processing systems}}  \bibinfo{volume}{27} (\bibinfo{year}{2014}).
\newblock


\bibitem[{Google Brain Team}(2016)]%
        {tfplayground}
\bibfield{author}{\bibinfo{person}{{Google Brain Team}}.} \bibinfo{year}{2016}\natexlab{}.
\newblock \bibinfo{title}{TensorFlow Playground}.
\newblock \bibinfo{howpublished}{\url{https://playground.tensorflow.org/}}.
\newblock
\newblock
\shownote{Accessed: July 2025}.


\bibitem[Hart and Staveland(1988)]%
        {hart1988development}
\bibfield{author}{\bibinfo{person}{Sandra~G. Hart} {and} \bibinfo{person}{Lowell~E. Staveland}.} \bibinfo{year}{1988}\natexlab{}.
\newblock \showarticletitle{Development of {NASA-TLX} (Task Load Index): Results of empirical and theoretical research}.
\newblock In \bibinfo{booktitle}{\emph{Advances in Psychology}}. Vol.~\bibinfo{volume}{52}. \bibinfo{publisher}{Elsevier}, \bibinfo{pages}{139--183}.
\newblock


\bibitem[He et~al\mbox{.}(2016)]%
        {he2016resnet}
\bibfield{author}{\bibinfo{person}{Kaiming He} {et~al\mbox{.}}} \bibinfo{year}{2016}\natexlab{}.
\newblock \showarticletitle{Deep residual learning for image recognition}. In \bibinfo{booktitle}{\emph{CVPR}}.
\newblock


\bibitem[Hohman et~al\mbox{.}(2019)]%
        {hohman2019survey}
\bibfield{author}{\bibinfo{person}{Fred Hohman}, \bibinfo{person}{Minsuk Kahng}, \bibinfo{person}{Robert Pienta}, {and} \bibinfo{person}{Duen~Horng Chau}.} \bibinfo{year}{2019}\natexlab{}.
\newblock \showarticletitle{Visual Analytics in Deep Learning: An Interrogative Survey for the Next Frontiers}.
\newblock \bibinfo{journal}{\emph{IEEE Transactions on Visualization and Computer Graphics}} \bibinfo{volume}{25}, \bibinfo{number}{8} (\bibinfo{year}{2019}), \bibinfo{pages}{2674--2693}.
\newblock
\urldef\tempurl%
\url{https://doi.org/10.1109/TVCG.2018.2843369}
\showDOI{\tempurl}


\bibitem[Hohman et~al\mbox{.}(2020)]%
        {hohman2020summit}
\bibfield{author}{\bibinfo{person}{Fred Hohman}, \bibinfo{person}{Haekyu Park}, \bibinfo{person}{Caleb Robinson}, {and} \bibinfo{person}{Duen~Horng Chau}.} \bibinfo{year}{2020}\natexlab{}.
\newblock \showarticletitle{Summit: Scaling Deep Learning Interpretability by Visualizing Activation and Attribution Summarizations}. In \bibinfo{booktitle}{\emph{IEEE Visualization Conference (VIS)}}.
\newblock
\urldef\tempurl%
\url{https://par.nsf.gov/servlets/purl/10183516}
\showURL{%
\tempurl}


\bibitem[Jian et~al\mbox{.}(2000)]%
        {jian2000foundations}
\bibfield{author}{\bibinfo{person}{Jiun-Yin Jian}, \bibinfo{person}{Ann~M. Bisantz}, {and} \bibinfo{person}{Colin~G. Drury}.} \bibinfo{year}{2000}\natexlab{}.
\newblock \showarticletitle{Foundations for an empirically determined scale of trust in automated systems}.
\newblock \bibinfo{journal}{\emph{International Journal of Cognitive Ergonomics}} \bibinfo{volume}{4}, \bibinfo{number}{1} (\bibinfo{year}{2000}), \bibinfo{pages}{53--71}.
\newblock


\bibitem[Kahng et~al\mbox{.}(2017)]%
        {kahng2017activis}
\bibfield{author}{\bibinfo{person}{Minsuk Kahng}, \bibinfo{person}{Abigail Andrews}, \bibinfo{person}{Anuja Kalro}, {and} \bibinfo{person}{Duen~Horng Chau}.} \bibinfo{year}{2017}\natexlab{}.
\newblock \showarticletitle{ActiVis: Visual Exploration of Industry-Scale Deep Neural Network Models}. In \bibinfo{booktitle}{\emph{IEEE Transactions on Visualization and Computer Graphics}}. \bibinfo{pages}{1--10}.
\newblock


\bibitem[Krizhevsky et~al\mbox{.}(2012)]%
        {krizhevsky2012imagenet}
\bibfield{author}{\bibinfo{person}{Alex Krizhevsky}, \bibinfo{person}{Ilya Sutskever}, {and} \bibinfo{person}{Geoffrey Hinton}.} \bibinfo{year}{2012}\natexlab{}.
\newblock \showarticletitle{ImageNet classification with deep convolutional neural networks}. In \bibinfo{booktitle}{\emph{NeurIPS}}.
\newblock


\bibitem[Lazar et~al\mbox{.}(2017)]%
        {lazar2017research}
\bibfield{author}{\bibinfo{person}{Jonathan Lazar}, \bibinfo{person}{Jinjuan~Heidi Feng}, {and} \bibinfo{person}{Harry Hochheiser}.} \bibinfo{year}{2017}\natexlab{}.
\newblock \bibinfo{booktitle}{\emph{Research Methods in Human-Computer Interaction}}.
\newblock \bibinfo{publisher}{Morgan Kaufmann}.
\newblock


\bibitem[LeCun et~al\mbox{.}(2015)]%
        {lecun2015deep}
\bibfield{author}{\bibinfo{person}{Yann LeCun}, \bibinfo{person}{Yoshua Bengio}, {and} \bibinfo{person}{Geoffrey Hinton}.} \bibinfo{year}{2015}\natexlab{}.
\newblock \showarticletitle{Deep learning}.
\newblock \bibinfo{journal}{\emph{Nature}} (\bibinfo{year}{2015}).
\newblock


\bibitem[LeCun et~al\mbox{.}(1989)]%
        {lecun1989backpropagation}
\bibfield{author}{\bibinfo{person}{Yann LeCun}, \bibinfo{person}{Bernhard Boser}, \bibinfo{person}{John~S Denker}, \bibinfo{person}{Donnie Henderson}, \bibinfo{person}{Richard~E Howard}, \bibinfo{person}{Wayne Hubbard}, {and} \bibinfo{person}{Lawrence~D Jackel}.} \bibinfo{year}{1989}\natexlab{}.
\newblock \showarticletitle{{Backpropagation applied to handwritten zip code recognition}}.
\newblock \bibinfo{journal}{\emph{Neural computation}} \bibinfo{volume}{1}, \bibinfo{number}{4} (\bibinfo{year}{1989}), \bibinfo{pages}{541--551}.
\newblock


\bibitem[Lee and See(2004)]%
        {lee2004trust}
\bibfield{author}{\bibinfo{person}{John~D. Lee} {and} \bibinfo{person}{Katrina~A. See}.} \bibinfo{year}{2004}\natexlab{}.
\newblock \showarticletitle{Trust in automation: Designing for appropriate reliance}.
\newblock \bibinfo{journal}{\emph{Human Factors}} \bibinfo{volume}{46}, \bibinfo{number}{1} (\bibinfo{year}{2004}), \bibinfo{pages}{50--80}.
\newblock


\bibitem[Likert(1932)]%
        {likert1932technique}
\bibfield{author}{\bibinfo{person}{Rensis Likert}.} \bibinfo{year}{1932}\natexlab{}.
\newblock \showarticletitle{A Technique for the Measurement of Attitudes}.
\newblock \bibinfo{journal}{\emph{Archives of Psychology}} \bibinfo{volume}{22}, \bibinfo{number}{140} (\bibinfo{year}{1932}), \bibinfo{pages}{1--55}.
\newblock


\bibitem[Lipton(2018)]%
        {lipton2018mythos}
\bibfield{author}{\bibinfo{person}{Zachary Lipton}.} \bibinfo{year}{2018}\natexlab{}.
\newblock \showarticletitle{The mythos of model interpretability}.
\newblock \bibinfo{journal}{\emph{Queue}} (\bibinfo{year}{2018}).
\newblock


\bibitem[{Lutz Roeder}(2022)]%
        {netron}
\bibfield{author}{\bibinfo{person}{{Lutz Roeder}}.} \bibinfo{year}{2022}\natexlab{}.
\newblock \bibinfo{title}{Netron}.
\newblock \bibinfo{howpublished}{\url{https://netron.app/}}.
\newblock
\newblock
\shownote{Accessed: July 2025}.


\bibitem[Mersha et~al\mbox{.}(2024)]%
        {scienceDirect2024xai_survey}
\bibfield{author}{\bibinfo{person}{Melkamu Mersha}, \bibinfo{person}{Khang Lam}, \bibinfo{person}{Joseph Wood}, \bibinfo{person}{Ali~K. AlShami}, {and} \bibinfo{person}{Jugal Kalita}.} \bibinfo{year}{2024}\natexlab{}.
\newblock \showarticletitle{Explainable artificial intelligence: A survey of needs, techniques, applications, and future direction}.
\newblock \bibinfo{journal}{\emph{Neurocomputing vol. 599}} (\bibinfo{year}{2024}).
\newblock
\urldef\tempurl%
\url{https://www.sciencedirect.com/science/article/abs/pii/S0925231224008828}
\showURL{%
\tempurl}


\bibitem[Mittelstadt et~al\mbox{.}(2019)]%
        {mittelstadt2019explaining}
\bibfield{author}{\bibinfo{person}{Brent Mittelstadt} {et~al\mbox{.}}} \bibinfo{year}{2019}\natexlab{}.
\newblock \showarticletitle{Explaining explanations in AI}.
\newblock \bibinfo{journal}{\emph{FAT*}} (\bibinfo{year}{2019}).
\newblock


\bibitem[Olah et~al\mbox{.}(2018)]%
        {olah2018building}
\bibfield{author}{\bibinfo{person}{Chris Olah} {et~al\mbox{.}}} \bibinfo{year}{2018}\natexlab{}.
\newblock \showarticletitle{The building blocks of interpretability}.
\newblock \bibinfo{journal}{\emph{Distill}} (\bibinfo{year}{2018}).
\newblock


\bibitem[Olah et~al\mbox{.}(2020)]%
        {olah2020zoom}
\bibfield{author}{\bibinfo{person}{Chris Olah} {et~al\mbox{.}}} \bibinfo{year}{2020}\natexlab{}.
\newblock \showarticletitle{Zoom In: An introduction to circuits}.
\newblock \bibinfo{journal}{\emph{Distill}} (\bibinfo{year}{2020}).
\newblock


\bibitem[Olah et~al\mbox{.}(2017)]%
        {olah2017featureviz}
\bibfield{author}{\bibinfo{person}{Chris Olah}, \bibinfo{person}{Alexander Mordvintsev}, {and} \bibinfo{person}{Ludwig Schubert}.} \bibinfo{year}{2017}\natexlab{}.
\newblock \showarticletitle{Feature Visualization}.
\newblock \bibinfo{journal}{\emph{Distill}} \bibinfo{volume}{2}, \bibinfo{number}{11} (\bibinfo{year}{2017}).
\newblock
\urldef\tempurl%
\url{https://doi.org/10.23915/distill.00007}
\showDOI{\tempurl}


\bibitem[{Polo Club of Data Science}(2018)]%
        {ganlab}
\bibfield{author}{\bibinfo{person}{{Polo Club of Data Science}}.} \bibinfo{year}{2018}\natexlab{}.
\newblock \bibinfo{title}{GAN Lab}.
\newblock \bibinfo{howpublished}{\url{https://poloclub.github.io/ganlab/}}.
\newblock
\newblock
\shownote{Accessed: July 2025}.


\bibitem[{Polo Club of Data Science}(2020)]%
        {cnnexplainer}
\bibfield{author}{\bibinfo{person}{{Polo Club of Data Science}}.} \bibinfo{year}{2020}\natexlab{}.
\newblock \bibinfo{title}{CNN Explainer}.
\newblock \bibinfo{howpublished}{\url{https://poloclub.github.io/cnn-explainer/}}.
\newblock
\newblock
\shownote{Accessed: July 2025}.


\bibitem[{Polo Club of Data Science}(2023)]%
        {transformerexplainer}
\bibfield{author}{\bibinfo{person}{{Polo Club of Data Science}}.} \bibinfo{year}{2023}\natexlab{}.
\newblock \bibinfo{title}{Transformer Explainer: Interactive Visualization of Transformer Models}.
\newblock \bibinfo{howpublished}{\url{https://poloclub.github.io/transformer-explainer/}}.
\newblock
\newblock
\shownote{Accessed: 2026-02-10}.


\bibitem[Rauber et~al\mbox{.}(2017)]%
        {rauber2017hidden}
\bibfield{author}{\bibinfo{person}{Paulo~E. Rauber}, \bibinfo{person}{Samuel~G. Fadel}, \bibinfo{person}{Alexandre~X. Falc{\~a}o}, {and} \bibinfo{person}{Alexandru~C. Telea}.} \bibinfo{year}{2017}\natexlab{}.
\newblock \showarticletitle{Visualizing the Hidden Activity of Artificial Neural Networks}.
\newblock \bibinfo{journal}{\emph{IEEE Transactions on Visualization and Computer Graphics}} \bibinfo{volume}{23}, \bibinfo{number}{1} (\bibinfo{year}{2017}), \bibinfo{pages}{101--110}.
\newblock
\urldef\tempurl%
\url{https://doi.org/10.1109/TVCG.2016.2598838}
\showDOI{\tempurl}


\bibitem[Rawassizadeh(2025)]%
        {reza2025}
\bibfield{author}{\bibinfo{person}{Reza Rawassizadeh}.} \bibinfo{year}{2025}\natexlab{}.
\newblock \bibinfo{booktitle}{\emph{Machine learning and artificial intelligence: concepts, algorithms and models}}.
\newblock \bibinfo{publisher}{Reza Rawassizadeh}.
\newblock


\bibitem[Schmidhuber(2015)]%
        {schmidhuber2015deep}
\bibfield{author}{\bibinfo{person}{Jürgen Schmidhuber}.} \bibinfo{year}{2015}\natexlab{}.
\newblock \showarticletitle{Deep learning in neural networks: An overview}.
\newblock \bibinfo{journal}{\emph{Neural Networks}} (\bibinfo{year}{2015}).
\newblock


\bibitem[Shneiderman(1996)]%
        {shneiderman1996eyes}
\bibfield{author}{\bibinfo{person}{Ben Shneiderman}.} \bibinfo{year}{1996}\natexlab{}.
\newblock \showarticletitle{The Eyes Have It: A Task by Data Type Taxonomy for Information Visualizations}. In \bibinfo{booktitle}{\emph{Proceedings of the IEEE Symposium on Visual Languages}}. \bibinfo{publisher}{IEEE}, \bibinfo{pages}{336--343}.
\newblock


\bibitem[Strobelt et~al\mbox{.}(2018)]%
        {strobelt2018lstmvis}
\bibfield{author}{\bibinfo{person}{Hendrik Strobelt}, \bibinfo{person}{Sebastian Gehrmann}, \bibinfo{person}{Hanspeter Pfister}, {and} \bibinfo{person}{Alexander~M. Rush}.} \bibinfo{year}{2018}\natexlab{}.
\newblock \showarticletitle{LSTMVis: A Tool for Visual Analysis of Hidden State Dynamics in Recurrent Neural Networks}.
\newblock \bibinfo{journal}{\emph{IEEE Transactions on Visualization and Computer Graphics}} \bibinfo{volume}{24}, \bibinfo{number}{1} (\bibinfo{year}{2018}), \bibinfo{pages}{667--676}.
\newblock
\urldef\tempurl%
\url{https://doi.org/10.1109/TVCG.2017.2744158}
\showDOI{\tempurl}


\bibitem[Sweller(1994)]%
        {sweller1994cognitive}
\bibfield{author}{\bibinfo{person}{John Sweller}.} \bibinfo{year}{1994}\natexlab{}.
\newblock \showarticletitle{Cognitive load theory, learning difficulty, and instructional design}.
\newblock \bibinfo{journal}{\emph{Learning and Instruction}} \bibinfo{volume}{4}, \bibinfo{number}{4} (\bibinfo{year}{1994}), \bibinfo{pages}{295--312}.
\newblock


\bibitem[Vaswani et~al\mbox{.}(2017)]%
        {vaswani2017attention}
\bibfield{author}{\bibinfo{person}{Ashish Vaswani} {et~al\mbox{.}}} \bibinfo{year}{2017}\natexlab{}.
\newblock \showarticletitle{Attention is all you need}. In \bibinfo{booktitle}{\emph{NeurIPS}}.
\newblock


\bibitem[Wang et~al\mbox{.}(2018)]%
        {wang2018ganviz}
\bibfield{author}{\bibinfo{person}{Junpeng Wang}, \bibinfo{person}{Liang Gou}, \bibinfo{person}{Hao Yang}, {and} \bibinfo{person}{Han-Wei Shen}.} \bibinfo{year}{2018}\natexlab{}.
\newblock \showarticletitle{GANViz: A Visual Analytics Approach to Understand the Adversarial Game}.
\newblock \bibinfo{journal}{\emph{IEEE Transactions on Visualization and Computer Graphics}} \bibinfo{volume}{24}, \bibinfo{number}{6} (\bibinfo{year}{2018}), \bibinfo{pages}{1905--1917}.
\newblock


\bibitem[Wilcoxon(1945)]%
        {wilcoxon1945individual}
\bibfield{author}{\bibinfo{person}{Frank Wilcoxon}.} \bibinfo{year}{1945}\natexlab{}.
\newblock \showarticletitle{Individual comparisons by ranking methods}.
\newblock \bibinfo{journal}{\emph{Biometrics Bulletin}} \bibinfo{volume}{1}, \bibinfo{number}{6} (\bibinfo{year}{1945}), \bibinfo{pages}{80--83}.
\newblock


\bibitem[Wongsuphasawat et~al\mbox{.}(2018)]%
        {wongsuphasawat2018tfgraph}
\bibfield{author}{\bibinfo{person}{Kanit Wongsuphasawat}, \bibinfo{person}{Daniel Smilkov}, \bibinfo{person}{James Wexler}, \bibinfo{person}{Jimbo Wilson}, \bibinfo{person}{Dandelion Man{\'e}}, \bibinfo{person}{Doug Fritz}, \bibinfo{person}{Dilip Krishnan}, \bibinfo{person}{Fernanda~B. Vi{\'e}gas}, {and} \bibinfo{person}{Martin Wattenberg}.} \bibinfo{year}{2018}\natexlab{}.
\newblock \showarticletitle{Visualizing Dataflow Graphs of Deep Learning Models in TensorFlow}.
\newblock \bibinfo{journal}{\emph{IEEE Transactions on Visualization and Computer Graphics}} \bibinfo{volume}{24}, \bibinfo{number}{1} (\bibinfo{year}{2018}), \bibinfo{pages}{1--12}.
\newblock
\urldef\tempurl%
\url{https://doi.org/10.1109/TVCG.2017.2744878}
\showDOI{\tempurl}


\bibitem[Yang et~al\mbox{.}(2023)]%
        {yang2023xai_review}
\bibfield{author}{\bibinfo{person}{Wenli Yang}, \bibinfo{person}{Yuchen Wei}, \bibinfo{person}{Hanyu Wei}, \bibinfo{person}{Yanyu Chen}, \bibinfo{person}{Guan Huang}, \bibinfo{person}{Xiang Li}, \bibinfo{person}{Renjie Li}, \bibinfo{person}{Naimeng Yao}, \bibinfo{person}{Xinyi Wang}, \bibinfo{person}{Xiaotong Gu}, \bibinfo{person}{Muhammad~Bilal Amin}, {and} \bibinfo{person}{Byeong Kang}.} \bibinfo{year}{2023}\natexlab{}.
\newblock \showarticletitle{Survey on Explainable AI: From Approaches, Limitations and Applications Aspects}.
\newblock \bibinfo{journal}{\emph{Human-Centric Intelligent Systems}}  \bibinfo{volume}{3} (\bibinfo{year}{2023}), \bibinfo{pages}{161--188}.
\newblock
\urldef\tempurl%
\url{https://doi.org/10.1007/s44230-023-00038-y}
\showDOI{\tempurl}


\bibitem[Yosinski et~al\mbox{.}(2015)]%
        {yosinski2015deepviz}
\bibfield{author}{\bibinfo{person}{Jason Yosinski}, \bibinfo{person}{Jeff Clune}, \bibinfo{person}{Anh Nguyen}, \bibinfo{person}{Thomas Fuchs}, {and} \bibinfo{person}{Hod Lipson}.} \bibinfo{year}{2015}\natexlab{}.
\newblock \showarticletitle{Understanding Neural Networks Through Deep Visualization}.
\newblock \bibinfo{journal}{\emph{arXiv preprint arXiv:1506.06579}} (\bibinfo{year}{2015}).
\newblock
\urldef\tempurl%
\url{https://arxiv.org/abs/1506.06579}
\showURL{%
\tempurl}


\end{thebibliography}

\end{document}